\documentclass[11pt]{article}

\usepackage{graphicx}%
\usepackage{multirow}%
\usepackage{amsmath,amssymb,amsfonts}%
\usepackage{amsthm}%
\usepackage{mathrsfs}%
\usepackage[title]{appendix}%
\usepackage{xcolor}%
\usepackage{textcomp}%
\usepackage{manyfoot}%
\usepackage{booktabs}%
\usepackage{algorithm}%
\usepackage{algorithmicx}%
\usepackage{algpseudocode}%
\usepackage{listings}%
\usepackage{pdfpages}
\usepackage[letterpaper,top=2cm,bottom=2cm,left=3cm,right=3cm,marginparwidth=1.75cm]{geometry}
\usepackage{bigstrut}
\usepackage{xspace}
\usepackage{longtable}
\usepackage{caption}
\usepackage{subcaption}
\usepackage{adjustbox}
\usepackage[colorlinks=true, linkcolor=black, citecolor=blue]{hyperref}
\usepackage[superscript]{cite}
\usepackage{lineno}

\makeatletter
\renewcommand{\maketitle}{\bgroup\setlength{\parindent}{0pt}
\begin{flushleft}
  \textbf{\@title}

  \@author
\end{flushleft}\egroup
}
\makeatother

\begin{document}

\newcommand{\recall}{\text{Recall}\xspace}
\newcommand{\acc}{\text{ACC}\xspace}
\newcommand{\precision}{\text{Precision}\xspace}
\newcommand{\llm}{\text{LLM}\xspace}
\newcommand{\method}{InformGen\xspace}
\newcommand{\datasetname}{InformBench\xspace}

\title{InformGen: An AI Copilot for Accurate and Compliant Clinical Research Consent Document Generation}
\author{Zifeng Wang$^{1}$, Junyi Gao$^{2}$, Benjamin Danek$^{1}$, Brandon Theodorou$^{1}$, Ruba Shaik$^{3}$, Shivashankar Thati$^{1}$, Seunghyun Won$^{4}$, Jimeng~Sun$^{1,3\#}$\\
$^{1}$ School of Computing and Data Science, University of Illinois Urbana-Champaign, Urbana, IL, USA \\
$^2$ Centre for Medical Informatics, Usher Institute, University of Edinburgh, Edinburgh, UK \\
$^3$ Carle Illinois College of Medicine, University of Illinois Urbana-Champaign, Champaign, IL, USA \\
$^4$ Seoul National University Bundang Hospital, Gyeonggi, Republic of Korea \\
$^\#$Corresponding author. Emails: jimeng@illinois.edu
}

\maketitle



\abstract{Leveraging large language models (LLMs) to generate high-stakes documents, such as informed consent forms (ICFs), remains a significant challenge due to the extreme need for regulatory compliance and factual accuracy. Here, we present \method, an LLM-driven copilot for accurate and compliant ICF drafting by optimized knowledge document parsing and content generation, with humans in the loop. We further construct a benchmark dataset comprising protocols and ICFs from 900 clinical trials. Experimental results demonstrate that \method achieves near 100\% compliance with 18 core regulatory rules derived from FDA guidelines, outperforming a vanilla GPT-4o model by up to 30\%. Additionally, a user study with five annotators shows that \method, when integrated with manual intervention, attains over 90\% factual accuracy, significantly surpassing the vanilla GPT-4o model’s 57\%-82\%. Crucially, \method ensures traceability by providing inline citations to source protocols, enabling easy verification and maintaining the highest standards of factual integrity.
}

\newpage

\section*{Introduction}\label{sec:intro}

Informed consent forms (ICFs) constitute a fundamental pillar of ethical clinical research, functioning as the primary medium through which participants comprehend study procedures, risks, benefits, and their rights \cite{grady2015enduring}. Clinical trials necessitate multiple ICF iterations to accommodate site-specific requirements, institutional review board stipulations, and linguistic-policy adaptations for multinational studies \cite{nishimura2013improving}. The ICF development process requires a comprehensive analysis of clinical trial protocols, which often span tens to hundreds of pages. This resource-intensive endeavor significantly impacts clinical research efficiency, potentially delaying participant enrollment and extending study timelines \cite{koonrungsesomboon2015understanding}, while simultaneously compromising trial inclusivity through suboptimal ICF quality and accessibility \cite{velez2023consent}.

Large language models (LLMs), trained on vast textual corpora, initially emerged as tools for narrative text generation applications \cite{yuan2022wordcraft}. Their integration into medicine has advanced substantially, facilitating clinical text generation and summarization \cite{van2024adapted} while enabling patient-oriented conversational interfaces \cite{wan2024outpatient}. Within clinical trials research, investigators have adapted LLMs for protocol development tasks, including the formulation of eligibility criteria and generation of trial synopses \cite{wang2023autotrial,lin2024panacea}. Recent investigations adopted LLMs to simplify ICFs to facilitate their accessibility \cite{mirza2024using,ali2024bridging}. Besides, a particularly promising method is retrieval-augmented generation (RAG) \cite{lewis2020retrieval}, wherein LLMs dynamically incorporate reference documents during text generation. RAG enables LLMs to process extensive source documentation while producing outputs with explicit citations grounded in sources, thereby facilitating human review and verification~\cite{gao2023enabling}.

Despite the advancements, the application of LLMs for ICF drafting remains largely unexplored, presenting several critical challenges. First, regulatory compliance poses a major hurdle, as ICFs must adhere to FDA guidelines on structure and language~\cite{food2023informed}, as well as institution-specific policies that vary across research sites \cite{stanfordICF,ucsfICF}. Second, clinical trial documents demand exceptional information fidelity, requiring strict adherence to trial protocols while ensuring content remains verifiable by human reviewers, an aspect insufficiently examined in prior research. Third, there is a lack of dedicated evaluation frameworks for AI-generated regulatory documents. Traditional natural language processing (NLP) metrics, such as BLEU scores~\cite{papineni2002bleu}, primarily assess text overlap and semantic similarity but fail to capture critical ICF-specific quality dimensions. Besides, existing AI-driven approaches for ICFs predominantly focus on isolated aspects, such as readability~\cite{decker2023large,vaira2025evaluating,shi2025transforming}, rather than addressing the full spectrum of requirements for regulatory compliance and content reliability. In particular, a recent study found that none of the LLMs tested, including GPT-4 and Gemini, were capable of generating a legally compliant ICF and only covered fewer than one-third of the required risks, procedural details, and preparatory instructions~\cite{raimann2024evaluation}.

In this work, we make three key contributions: (1) \datasetname, a benchmark dataset comprising over 900 paired clinical trial protocols and their corresponding ICFs, representing the largest AI-ready ICF drafting dataset to date; (2) an evaluation framework aligned with FDA guidelines~\cite{food2023informed} to systematically assess compliance and factual accuracy, leveraging both automated and human evaluations for benchmarking LLM performance; and (3) \method, a specialized LLM pipeline that enhances human-AI collaboration in ICF generation. Our evaluation demonstrates that this approach achieves near-perfect regulatory compliance (in most cases 100\%), maintains high factual accuracy (80-95\%), and offers variable and editable evidence, within a human-in-the-loop framework.

\section*{Results}
\subsection*{Overview of \method}
Figure~\ref{fig:overview} presents an overview of the development and evaluation of the \method framework for informed consent form (ICF) drafting. The name \method reflects three aspects of ``informed" principles in this study. First, it is knowledge- and policy-informed, as it integrates input from clinical trial protocols, FDA regulations, and site-specific policies to guide document generation. Second, it is human-informed, enabling seamless verification and correction of both inputs and outputs within the pipeline. Finally, it is designed specifically as an AI copilot for informed consent form generation. Specifically, \method consists of three key stages: (1) knowledge document parsing, (2) document drafting, and (3) content evaluation (See ``Methods"). 

Knowledge document parsing processes two main types of documents prior to ICF generation as illustrated in Figure~\hyperref[fig:overview]{1a}. (see ``Methods"). The first type includes the \textit{guideline document} and \textit{consent template}, which provide regulatory guidelines and site-specific policies, respectively. The guideline document applies universally to all ICFs and requires parsing only once, while the consent template is site-specific and remains applicable to all future ICFs from the same site. These documents are parsed and transformed into structured rules that are attached to relevant ICF sections. The second type of document is the \textit{clinical trial protocol}, which is essential for each individual clinical trial and contains critical information required for drafting the ICF. Specifically, two procedure-specific components are extracted to inform trial participants: the trial schedule and the risks associated with experimental procedures. To ensure accuracy, we take an additional step to extract and standardize the \textit{Schedule of Assessments} (SOA) table and the \textit{procedure-specific risk} table for manual inspection before incorporating them into the ICF generation process. This step is necessary because a standard retrieval-augmented generation (RAG) approach often struggles to capture nuanced details when processing lengthy protocol documents spanning hundreds of pages.

Figure~\hyperref[fig:overview]{1b} illustrates the document drafting pipeline, which follows a RAG paradigm to ensure that all generated content is grounded in authoritative knowledge documents. Each generated section includes citation links indicating the exact source paragraph and section, enabling human verification. Aligned with site-specific policies, the pipeline incorporates section-level guidelines as directives and retrieves relevant reference text chunks to generate the corresponding ICF sections. For key sections such as \textit{``Procedures"}, the pipeline integrates the \textit{Schedule of Assessments} (SOA) table as an additional input, while for \textit{``Risks and Discomforts"}, it incorporates the procedure-specific risk table to enhance accuracy and compliance.

\subsection*{Overview of \datasetname}
Figure~\hyperref[fig:overview]{1c} illustrates the content evaluation framework, which assesses the generated content along two key dimensions. First, \textit{compliance} is evaluated by measuring the proportion of regulatory and site-specific rules that are correctly followed. We provide the parsed rules in Supplementary Figure~\ref{fig:fda_policy}. Second, \textit{factual accuracy} is assessed by calculating the ratio of key facts correctly covered in the generated content, based on reference human-written ICFs and the corresponding trial protocols.

\begin{figure}
    \centering
    \includegraphics[width=0.95\linewidth]{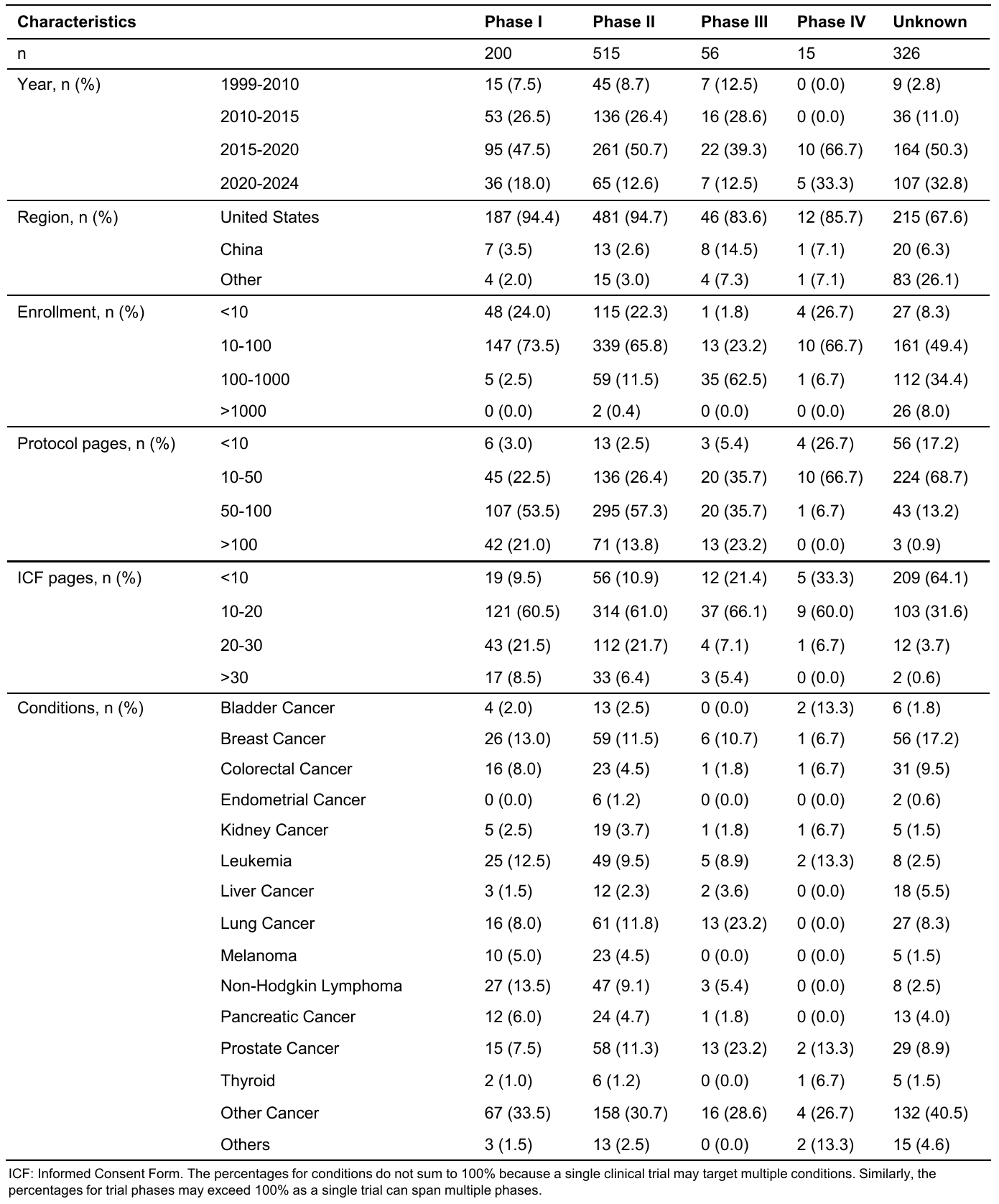}
    \caption{\textbf{Dataset statistics}. Characteristics of clinical trials and the associated protocols and informed consent forms used in the experiments. Year: the start year of the clinical trial; Region: which region this trial was primarily initiated and conducted; Enrollment: the number of actual enrollment subjects in the trial; Protocol pages: the number of pages of the clinical trial protocol; ICF pages: the number of pages of the human written consent form document; Conditions: the primary targeted conditions of the trial.}
    \label{fig:statistics}
\end{figure}

To comprehensively evaluate both generic LLMs and the optimized \method framework in terms of compliance and factual accuracy, we constructed the \datasetname benchmark. The dataset was built by first retrieving all available clinical trial protocols and ICFs from ClinicalTrials.gov. Among these, oncology-related ICFs exhibited higher quality in terms of formatting and adherence to regulatory standards. Consequently, we selected a subset primarily focused on oncology-related conditions, yielding a total of 900 trials with at least one available ICF and corresponding protocol document. The dataset statistics are presented in Figure~\ref{fig:statistics}. The included trials span all common cancer types~\cite{commoncancer} and all four clinical trial phases. Most protocol documents range from 50 to 100 pages, while ICFs typically contain 10 to 20 pages. A small subset of lower-quality protocols, often associated with trials labeled as having an ``Unknown" phase, contain fewer than 10 pages. In subsequent sections, we analyze how the quality and complexity of input protocols impact the quality of ICF generation.

Guided by an ICF template from Stanford~\cite{stanfordICF}, we focused on four critical and challenging sections: \textit{``Purpose of Research"}, \textit{``Procedures"}, \textit{``Duration of Study Involvement"}, and \textit{``Risks and Discomforts"}. Content quality was assessed specifically for these sections. Given the diverse template formats across ICFs, we applied a standardization approach to categorize each parsed section into one of the four sections or mark it as not applicable.

\subsection*{Evaluation of compliance}
\begin{figure}
    \centering
    \includegraphics[width=0.95\linewidth]{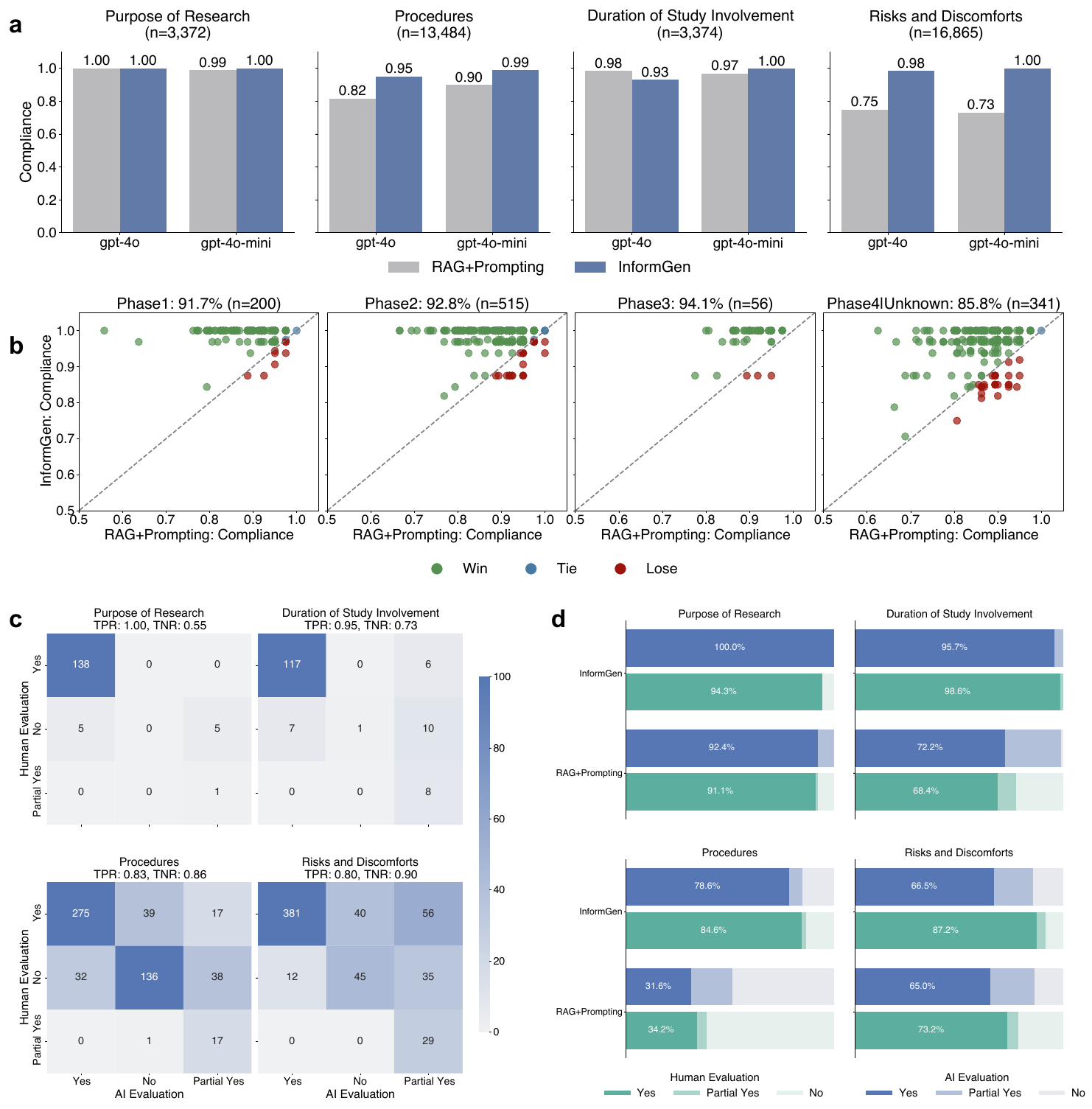}
    \caption{\textbf{Evaluating compliance of \method and baselines}. \textbf{a}, Automatic evaluation of the compliance rate of the generated content by \method and the baseline RAG+Prompting, using GPT-4o as a judge. Both methods are evaluated in two variants, utilizing either GPT-4o or GPT-4o-mini as the underlying LLM. \textbf{b}, In-depth trial-level comparison of \method and RAG+Prompting, analyzing compliance rates across all four clinical trial phases. \textbf{c}, Confusion matrix comparing compliance judgments made by GPT-4o and human annotators for the generated ICF content. TPR: True Positive Rate; TNR: True Negative Rate. \textbf{d}, Manual evaluation and GPT-4o-based assessment of compliance for the generated content across 100 trials.
}
    \label{fig:rule_validation}
\end{figure}

Figure~\ref{fig:rule_validation} presents the compliance evaluation results of \method and baseline models. As previously discussed, we focused on four key ICF sections, each associated with multiple regulatory rules (Supplementary Figure~\ref{fig:fda_policy}). To assess compliance, we conducted an automatic evaluation using GPT-4o on all ICFs, complemented by a manual evaluation on a subset of 100 trials. For each section in each trial, we evaluated whether the generated content adhered to each applicable rule, categorizing it as either met or violated. The compliance score was then computed as the proportion of rules successfully met. In the human evaluation, we introduced an intermediate category, ``Partial Yes", allowing annotators to indicate cases where a rule was only partially satisfied. For comparison, we implemented a baseline method that follows a standard RAG approach, supplemented with general guidance on drafting the target section (referred to as ``prompt"). We denote this approach as \textit{RAG+Prompting}. Notably, both \method and RAG+Prompting utilize the same underlying LLM, either GPT-4o or GPT-4o-mini, to ensure a fair comparison.

Figure~\hyperref[fig:rule_validation]{3a} presents the automatic evaluation results comparing \method and the baseline. We tested two variants of each method using GPT-4o and GPT-4o-mini, where the former represents a state-of-the-art LLM and the latter a lightweight version. The results indicate that \method, with explicit regulatory guidance and site-specific policies, achieved near 100\% compliance in most cases, regardless of the underlying LLM. In contrast, the {RAG+Prompting} baseline met compliance requirements only for the \textit{``Purpose of Research"} and \textit{``Duration of Study Involvement"} sections. The rules governing these sections are relatively straightforward, such as clearly stating the purpose of the clinical study and specifying its expected duration (Supplementary Figure~\ref{fig:fda_policy}). However, RAG+Prompting exhibited lower compliance rates for the \textit{``Procedures"} section (82\%-90\%) and the \textit{``Risks and Discomforts"} section (73\%-75\%).  For \textit{``Procedures"}, RAG+Prompting often failed to explicitly distinguish experimental procedures unique to the research, a key regulatory requirement. In the \textit{``Risks and Discomforts"} section, failure modes were more varied, reflecting the complexity of accurately identifying risks such as potential privacy loss, reasonably foreseeable discomforts, and the need to balance comprehensiveness without overwhelming the reader (Supplementary Figure~\ref{fig:fda_policy}). These challenges highlight the advantages of an optimized \method framework in mitigating such issues and ensuring higher compliance.

Figure~\hyperref[fig:rule_validation]{3b} provides a trial-level comparison of \method and \textit{RAG+Prompting}, further broken down by clinical trial phases. Across all phases, \method consistently outperforms the baseline, achieving a winning rate of over 90\% in most cases. Additionally, \method maintains a compliance rate exceeding 90\% across the majority of trials, reinforcing the overall trend observed in the average performance analysis.

Figure~\hyperref[fig:rule_validation]{3c} presents the confusion matrix comparing the alignment between automatic evaluation using GPT-4o and human evaluation. This analysis validates the reliability of using GPT-4o as an automated compliance judge. To ensure consistency with human annotators, GPT-4o was tasked with assessing rules using the same categories: ``Yes", ``No", or ``Partial Yes". Human assessments were treated as the ground truth. The results indicate that GPT-4o tended to be more stringent than human annotators, particularly for the \textit{``Purpose of Research"} and \textit{``Duration of Study Involvement"} sections, where the true negative rate (TNR) was noticeably lower than the true positive rate (TPR). For \textit{``Procedures"} and \textit{``Risks and Discomforts"}, GPT-4o also demonstrated strong agreement with human evaluations, achieving a TPR above 80\% and a TNR exceeding 85\%. These findings suggest that GPT-4o serves as a reliable compliance judge. Importantly, since this evaluation method does not require a ground truth ICF, it can be applied as a post-checking mechanism for any newly drafted ICF to ensure maximum compliance.

Figure~\hyperref[fig:rule_validation]{3d} compares the compliance performance of \method and RAG+Prompting across four key ICF sections, evaluated by both human annotators and GPT-4o on 100 clinical trials. \method consistently outperforms the baseline, achieving near-perfect compliance in \textit{``Purpose of Research"} and \textit{``Duration of Study Involvement"} according to both evaluations. In contrast, {RAG+Prompting} shows noticeably lower compliance, particularly in the \textit{``Duration of Study Involvement"} section. The gap between the methods is even more pronounced in the \textit{``Procedures"} and \textit{``Risks and Discomforts"} sections, where \method maintains high compliance, while {RAG+Prompting} struggles significantly, with compliance rates nearly halved. These results underscore \method’s effectiveness in adhering to regulatory requirements, particularly in sections requiring precise procedural and risk-related details.

\subsection*{Evaluation of factual accuracy}
Research consent forms impose stringent requirements on the accuracy of the information conveyed to clinical trial participants, with critical legal and ethical implications~\cite{allen2024augmenting}. However, LLMs have been observed to occasionally fabricate or omit essential trial-specific details, such as procedural risks and study schedules, because of the complexity and volume of the input knowledge documents like protocols. As such, we establish an evaluation framework for factual accuracy. For each ICF section, we first extract up to five key facts from human-written reference ICFs. The generated content is then assessed against these reference facts to determine whether each key fact is correctly represented (Figure~\hyperref[fig:overview]{1c}). The extracted key facts are tailored to the specifics of each section. For instance, in \textit{``Duration of Study Involvement"}, key facts include the total study duration in months and the expected number of participant visits. This approach allows us to systematically measure the factual accuracy of the generated content.

\begin{figure}
    \centering
    \includegraphics[width=0.95\linewidth]{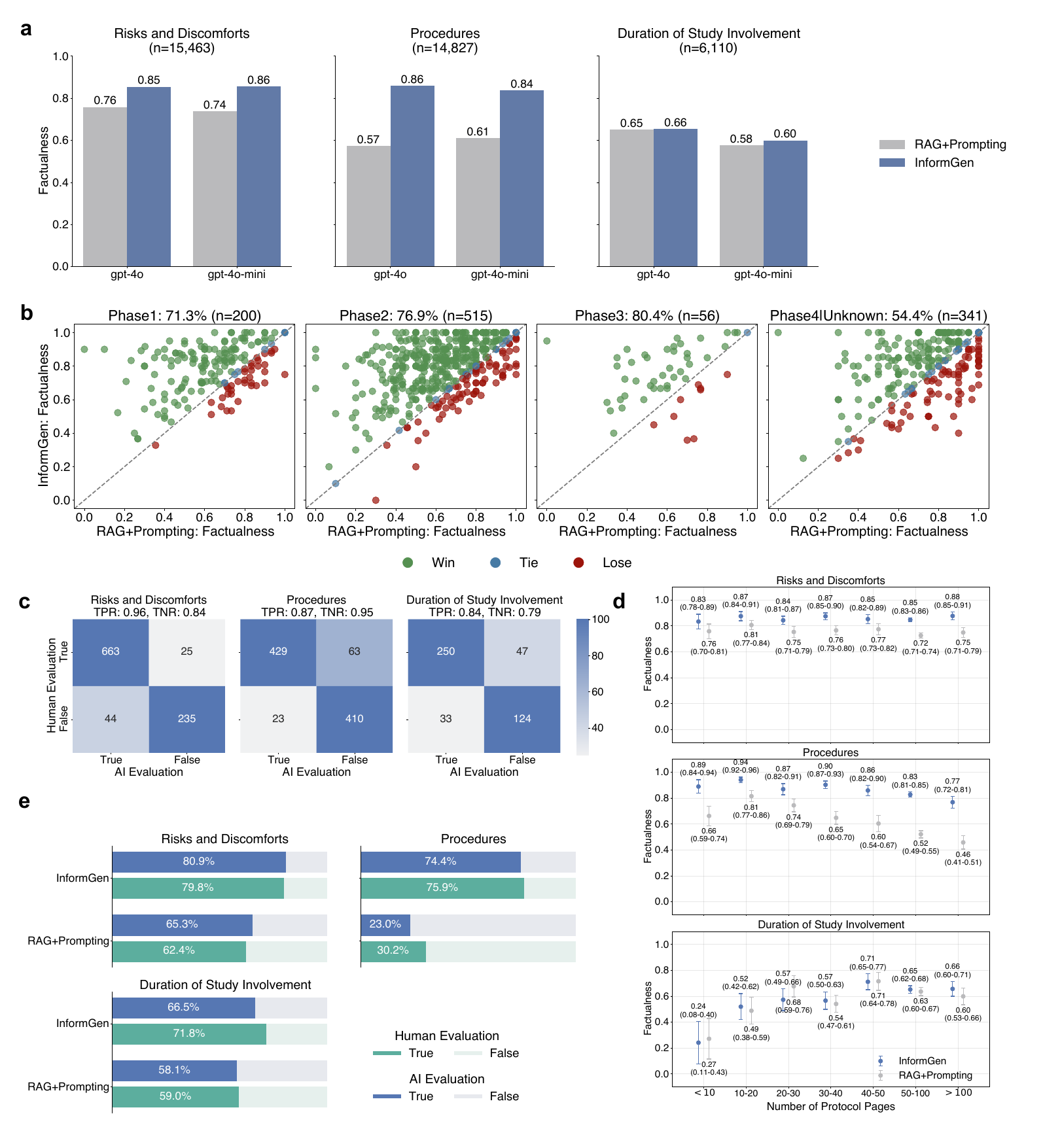}
    \caption{\textbf{Evaluating the factual accuracy of \method and baselines}. \textbf{a}, Automatic evaluation of the factual accuracy of content generated by \method and the baseline RAG+Prompting. Each ICF section is assessed against 3$\sim$5 key facts, with $n$ indicating the total number of evaluated facts. \textbf{b}, Trial-level breakdown of results by clinical trial phase. A side-by-side comparison of \method and RAG+Prompting for each trial shows that \method achieves a winning rate of 55\%-80\% in factual accuracy improvement. \textbf{c}, Confusion matrix comparing factual accuracy evaluations from GPT-4o (automatic evaluation) with human annotators. TPR: True Positive Rate; TNR: True Negative Rate. \textbf{d}, Analysis of factual accuracy across trials with varying protocol complexity. The x-axis represents the number of protocol pages, showing that \method exhibits greater robustness compared to RAG+Prompting as protocol complexity increases. \textbf{e}, Comparison of manual evaluation and GPT-4o-based assessment of factual accuracy across 100 trials.}
    \label{fig:fact_validation}
\end{figure}

Figure~\hyperref[fig:fact_validation]{4a} presents the factual accuracy results of \method and {RAG+Prompting} across three key ICF sections: \textit{``Risks and Discomforts"}, \textit{``Procedures"}, and \textit{``Duration of Study Involvement"}. The results represent the average accuracy per key fact across 900 clinical trials. \method achieves over 85\% factual accuracy in the \textit{``Risks and Discomforts"} and \textit{``Procedures"} sections, demonstrating a significant improvement over {RAG+Prompting} by 10\% and 30\% in absolute accuracy, respectively. This highlights the effectiveness of incorporating the SOA table and procedure-specific risk table extraction. In contrast, {RAG+Prompting} often omits crucial details in risk and procedure descriptions due to retrieval limitations. For the \textit{``Duration of Study Involvement"} section, the factual accuracy difference between the two methods is less pronounced. Analysis reveals that errors primarily stem from ambiguities in defining the expected duration of active participation, even when using the SOA table. Additionally, inaccuracies introduced during protocol document parsing further contribute to errors. These findings underscore the importance of incorporating a human-in-the-loop approach to enhance factual accuracy.

A similar pattern emerges in the trial-level factual accuracy comparison (Figure~\hyperref[fig:fact_validation]{4b}), where \method achieves a winning rate of 70\%-80\% for Phase 1 to Phase 3 trials. An exception is observed in \textit{``Phase 4 or Unknown"} trials, where \method attains a lower winning rate of 54.4\% over {RAG+Prompting}. Further analysis indicates that protocol quality negatively impacts the results, as 17.2\% of \textit{``Unknown"} phase trial protocols contain fewer than 10 pages, and 68.7\% have fewer than 50 pages (Figure~\ref{fig:statistics}). This trend is further validated by the factual accuracy analysis based on protocol length (Figure~\hyperref[fig:fact_validation]{4d}), which reveals a significant drop in accuracy for the \textit{``Duration of Study Involvement"} section when protocol documents contain fewer than 10 pages. Additionally, Figure~\hyperref[fig:fact_validation]{4d} highlights a general decline in performance as protocol length increases, though {RAG+Prompting} exhibits a steeper drop, while \method remains more stable. This discrepancy arises because vanilla retrieval approaches struggle to extract complete information from longer, more complex knowledge documents, whereas \method is more resilient to document length variations.

Figure~\hyperref[fig:fact_validation]{4c} illustrates the agreement between human annotators and the GPT-4o judge (AI evaluation) in the fact-checking process. Treating human evaluation as the ground truth, we calculate the true positive rate (TPR) and true negative rate (TNR) for AI evaluations. The results show that AI evaluation achieves a TPR of 0.84–0.96 and a TNR of 0.79–0.95, demonstrating strong alignment with human assessments. Additionally, we compare the absolute factual accuracy between human and AI evaluations for 100 clinical trials (Figure~\hyperref[fig:fact_validation]{4e}). The overall accuracy is similar between the two evaluation methods, though AI evaluation tends to be more stringent. \method consistently outperforms RAG+Prompting across all three sections, with the largest performance gap observed in the \textit{``Procedures"} section, where \method achieves 75\% accuracy compared to only 30\% for the baseline. These results highlight the effectiveness of the optimized \method pipeline in enhancing factual accuracy for generated content.

\subsection*{Human intervention improves compliance and factual accuracy}
Previous experiments have demonstrated the superiority of \method over the standard RAG-plus-prompting approach in ensuring regulatory compliance and improving factual accuracy in ICF generation. While compliance issues have been effectively addressed, a gap remains in achieving perfect factual accuracy, particularly in the \textit{``Duration of Study Involvement"} section, representing the ``last mile" for making this tool fully practical. Notably, all prior experiments were conducted under a fully automated setup, without human intervention. In real-world applications, human experts can be incorporated into the process to review and correct errors, further enhancing accuracy. To explore this potential, we have developed a demo platform that integrates human oversight, as illustrated in Figure~\ref{fig:manual_intervention}, and evaluated to what extent the lift it can bring.

Figure~\hyperref[fig:manual_intervention]{5a} presents example parsing results for two critical knowledge documents: the procedure-specific risk table and the schedule of assessment (SOA) table. Before \method proceeds with content generation, human users can review these intermediate outputs to ensure accuracy. Each procedure-risk pair, along with its frequency description, is linked to the corresponding source pages in the trial protocols, allowing users to verify the original context and make necessary edits. Similarly, the SOA table is standardized into a machine-readable format and linked to the extracted table from the protocol for detailed inspection. Once approved, these documents can be consistently applied across all ICFs for the specific trial. As illustrated in Figure~\hyperref[fig:manual_intervention]{5b}, the generated ICF content includes inline citations that link directly to the source content, with key references highlighted to indicate their role in LLM-based generation. This feature enables users to efficiently verify content against source documents and make necessary refinements. Additionally, users can leverage the AI copilot to assist in rewriting or refining the initial draft, ensuring both accuracy and clarity.

Figure~\hyperref[fig:manual_intervention]{5c} presents the manually evaluated factual accuracy results for 20 clinical trials, comparing two methods: RAG+Prompting and \method+Human. The former represents the baseline approach, while the latter incorporates human intervention to refine the SOA and procedure-risk tables before generating content for the three key ICF sections. The results demonstrate the effectiveness of human-in-the-loop refinement, with \method achieving up to 90\% factual accuracy when using GPT-4o as the underlying LLM. This represents a substantial absolute accuracy improvement of 15\% to 30\%, with high statistical significance. Moreover, errors in the generated content can be directly traced back to the extracted tables, allowing for efficient post-processing corrections. In contrast, the standard RAG approach struggles to identify missing content, often requiring a full proofread of the entire protocol. Additionally, we observed that GPT-4o-mini has difficulty accurately calculating duration-related values, suggesting that a more powerful base LLM is preferable for optimal performance.

\section*{Discussion}
Informed consent forms (ICFs) are essential documents for participant enrollment and onboarding in clinical research, subject to strict compliance and factual accuracy requirements to uphold ethical standards. Large language models (LLMs) have been utilized for drafting clinical trial documents~\cite{wang2023autotrial,lin2024panacea,markey2025rags} and for simplifying ICFs to enhance accessibility~\cite{mirza2024using,ali2024bridging}. However, the application of LLMs for ICF drafting remains largely unexplored, and no standardized framework exists for rigorously evaluating compliance and factual accuracy in LLM-generated clinical trial documents, particularly in ICFs. To address this gap, we constructed a dataset comprising protocols and ICFs from 900 clinical trials. We systematically parsed these documents to develop a benchmark dataset, \datasetname, along with a comprehensive evaluation framework to assess both compliance with regulatory standards and factual accuracy in LLM-generated ICF content. This evaluation framework can be executed by human experts but also by LLM as a judge, which, through our experiments, is highly consistent with how human experts' preferences.

On \datasetname, we implemented LLM-based ICF drafting using GPT-4o and GPT-4o-mini, following the standard approach of retrieval-augmented generation (RAG)~\cite{lewis2020retrieval} combined with prompt engineering~\cite{NEURIPS2020_1457c0d6}. Our experiments indicate that this RAG+Prompt method, even when leveraging state-of-the-art LLMs, falls short in meeting compliance and factual accuracy requirements, highlighting significant challenges in adopting LLMs for high-stakes document drafting in clinical trials. To address these limitations, we developed \method, an optimized ICF drafting pipeline designed to enhance both compliance and factual accuracy while incorporating human oversight. We compared \method against RAG+Prompting using the same underlying LLMs on \datasetname. The results demonstrate the superiority of \method: it achieved near 100\% compliance, ensuring adherence to regulatory requirements for language and content completeness. Additionally, \method outperformed RAG+Prompting in factual accuracy, with an absolute improvement of 10\%-20\% in the \textit{``Risks and Discomforts"} and \textit{``Procedures"} sections.

In a fully automated generation setup, we observed that \method did not show a significant improvement in the \textit{``Duration of Study Involvement"} section. To address this, we developed a demo platform that enables human interaction with \method, allowing users to review and refine intermediate knowledge documents, including the procedure-specific risk table and the schedule of assessment (SOA) table extracted from clinical trial protocols. Using this platform, we conducted a user study on 20 clinical trials, where participants could verify and modify these extracted documents before ICF generation. The results demonstrated a substantial benefit from human intervention, with \method achieving an absolute accuracy improvement of 20\%-30\% and reaching an overall accuracy of 80\%-95\%, including significant gains in the study duration section. This has highlighted the benefit and necessity in enhancing human-AI interaction when drafting high-stakes clinical trial documents such as ICFs.

This study has several limitations. (1) While \method is designed for clinical researchers, enhancing the experience for research participants remains an important consideration, particularly in addressing linguistic and regional diversity~\cite{velez2023consent}. (2) Although \method focuses on consent documents in clinical research, its application could be expanded to other clinical research documents and consent forms beyond clinical trials, such as surgical consent forms~\cite{ali2024bridging}, which remain largely unexplored. (3) The use of generic LLMs raises privacy concerns, as it requires data sharing with third-party model providers. Recent advancements in developing specialized LLMs based on open-source models~\cite{wang2025foundation} present a promising direction for improving security while leveraging clinical data to enhance model performance. (4) A larger-scale user study is needed to assess not only compliance and factual accuracy but also the practical benefits, including real-time efficiency gains in clinical workflows.

LLMs have made significant advancements in medical AI applications. \method highlights the critical need for adapting generic LLMs to clinical research while ensuring regulatory compliance and factual accuracy, while also incorporating human expertise in the loop. We anticipate that this paradigm will drive further research and expand LLM applications in clinical research.

\section*{Methods}

\subsection*{Collection and Building of \datasetname}
To construct \datasetname, we retrieved the full ClinicalTrials.gov database from the AACT database~\cite{aactdatabase}, which contains over 500,000 clinical trials. We first applied filters to identify oncology-related trials, starting with common cancer types~\cite{commoncancer} and using GPT-4o to generate an expanded set of keywords for each type. These keywords were then used to match target conditions in the ``conditions" table. Next, we extracted trial-provided documents, including protocols and informed consent forms, from the ``provided\textunderscore documents" table. Since not all trials included these documents, we retained 900 oncology trials with available materials. All retrieved documents were in PDF format and were converted into machine-readable markdown format using AWS Textract~\cite{awstextract}, an optical character recognition (OCR) service. Textract also facilitated section-wise segmentation of ICF and protocol content, which we leveraged to structure the dataset into samples with paired input protocol content and the output sections of ICFs.

The collected ICFs exhibit diverse structures, as they follow different templates and vary in quality. To enable a fair comparison across clinical trials, we standardized section classification. Specifically, for each parsed raw section, we utilized GPT-4o-mini to classify it into one of the predefined target sections. The target sections were determined based on the most critical components of the Stanford ICF template~\cite{stanfordICF}, including \textit{``Purpose of Research"}, \textit{``Procedures"}, \textit{``Duration of Study Involvement"}, and \textit{``Risks and Discomforts"}. Sections classified under the same target category were then merged to form a standardized reference ICF section, which serves as the basis for evaluation.

FDA establishes mandatory regulatory requirements for all ICFs. To systematically evaluate compliance, we developed a set of rules based on the FDA guidelines for informed consent~\cite{food2023informed}, specifically focusing on the \textit{``Basic Elements of Informed Consent"} section. This section outlines the essential components that an ICF must include, along with their intended purpose and explanations. We extracted and structured these components into a rule table, where each rule is mapped to a corresponding standard ICF section, enabling its use as a reference for compliance evaluation. The complete rule table is presented in Supplementary Figure~\ref{fig:fda_policy}. To assess content compliance, we provide the rule name and description as input to the GPT-4o judge, prompting it to determine whether each rule is followed or violated, producing a binary compliance assessment.

To generate labels for factual accuracy evaluation, we leveraged the ICF template, which outlines the key elements that must be included in each section. For example, in the \textit{``Purpose of Research"} section, the template specifies language such as: \textit{``This research study is looking for [state number of people] with [disease or condition]. [Clarify if enrollment will occur throughout the United States or internationally]. Stanford University expects to enroll [state number] research study participants."} Based on these guidelines, we developed section-specific criteria for extracting key information. In this case, relevant elements include the total number of participants, target conditions, and primary trial locations. We then provided these structured extraction guidelines to GPT-4o, enabling large-scale automated extraction across all sections of the 900 clinical trials in the dataset.

\subsection*{Clinical trial protocol parsing}
Before drafting a site-specific ICF for a trial, the first step is to comprehend the clinical trial protocol, as it serves as the foundation for study design, treatment schedules, and potential risks. Clinical trial protocols typically span tens to hundreds of pages, making it time-consuming even for experienced experts to extract relevant information for ICF drafting, as key details are often dispersed throughout the document. To address this challenge, we adopt a retrieval-augmented generation (RAG) approach, beginning by embedding the protocol into a vector database. This enables efficient retrieval of relevant information, ensuring that ICF drafting is grounded in the most pertinent trial details.

To process a protocol PDF, we first utilized Textract~\cite{awstextract}, an OCR service from AWS, to convert the raw file into Markdown, a machine-readable format that includes basic annotations of the page and section structure. Next, we segmented the Markdown file into text chunks using LangChain's \texttt{MarkdownHeaderTextSplitter}~\cite{langchainmdsplitter}, associating each chunk with its corresponding first- and second-level headers and page numbers as metadata. We then applied OpenAI's embedding model, \texttt{text-embedding-3-small}~\cite{openaiembedding}, to convert each text chunk into a 128-dimensional embedding vector. These embeddings were stored in a vector database, with Amazon MemoryDB as the primary implementation, though the approach is compatible with other open-source alternatives. This setup enables real-time retrieval, allowing any user query to be transformed into embeddings and matched against the most relevant protocol chunks efficiently.

Under the \method framework, two specialized knowledge documents require parsing: the schedule of assessment (SOA) table and the procedure-risk table. For SOA extraction, we developed an LLM-driven pipeline that begins by analyzing the raw protocol PDF to identify pages containing the SOA table. The selected pages, along with their parsed Markdown content, are then processed using GPT-4o with vision capabilities, following a structured extraction process: (1) detecting key procedures and time points, (2) extracting procedure-timepoint pairs along with corresponding activities, and (3) reconstructing the SOA table into a machine-readable format, such as an Excel spreadsheet, for user review. For the procedure-risk table, a similar pipeline is applied: (1) identifying pages mentioning experimental procedures, (2) extracting associated risks and discomforts for each procedure, including expected likelihood and relevant notes, and (3) reconstructing the procedure-risk table for user verification. This structured approach ensures accurate extraction and facilitates human oversight before integrating these documents into ICF drafting. The example output SOA table and procedure-risk table can be found in Figure~\hyperref[fig:manual_intervention]{5a}.

\subsection*{Document drafting}
\begin{figure}
    \centering
    \includegraphics[width=0.95\linewidth]{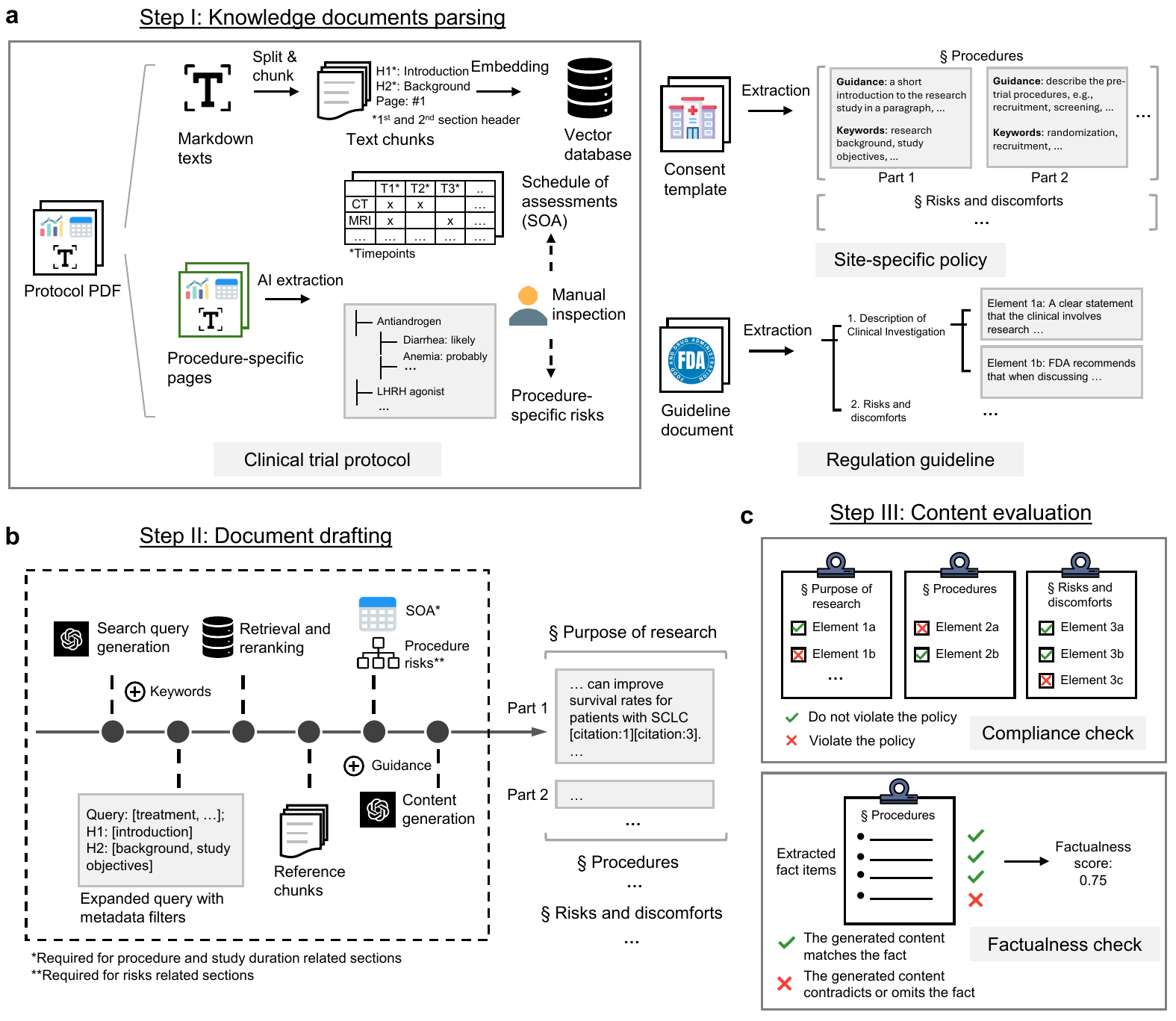}
    \caption{\textbf{Overview of \method Workflow}. \textbf{a}, The clinical trial protocol is parsed as the primary knowledge document. The raw PDF is converted to Markdown text, segmented into chunks with metadata on section titles and page numbers, and encoded into dense embeddings for storage in a vector database. Additionally, two specialized knowledge documents, the schedule of assessment (SOA) table and the procedure-risk table, are extracted from protocols and subjected to human review before use in content generation to ensure high factual accuracy. On the right, the FDA guidelines for ICF drafting are processed to create regulatory policies for compliance evaluation, while the site-specific consent template is parsed to provide structured instructions for content generation.  \textbf{b}, \method generates ICF sections using a retrieval-augmented generation pipeline. It first formulates search queries with metadata filters, retrieves the most relevant chunks, and applies parsed instructions to generate the content. For sections requiring trial schedules or risk-related details, the SOA or procedure-risk table is incorporated as an additional input. \textbf{c}, The generated content is evaluated for regulatory compliance and factual accuracy. Section-specific rule sets, derived from FDA guidelines, are used to assess whether the content violates any compliance requirements. Additionally, key facts are extracted from human-written reference ICFs as ground truth and compared against the generated content to compute an accuracy score.}
    \label{fig:overview}
\end{figure}

\method is designed to generate ICFs section by section while adhering to site-specific templates. To achieve this, we first parse the site templates into a structured format to provide clear guidance for LLMs. For each section in the template, we extract the content guidelines and, if necessary, break them down into multiple steps for better clarity. Each section is then associated with multiple content guidelines, specifying what should be included in the section. Additionally, we extract relevant keywords from these guidelines to generate search queries for retrieving pertinent information from the vector database (Figure~\hyperref[fig:overview]{1a}). This structured approach ensures that \method produces ICF content that aligns with site-specific requirements.

\begin{figure}
    \centering
    \includegraphics[width=0.95\linewidth]{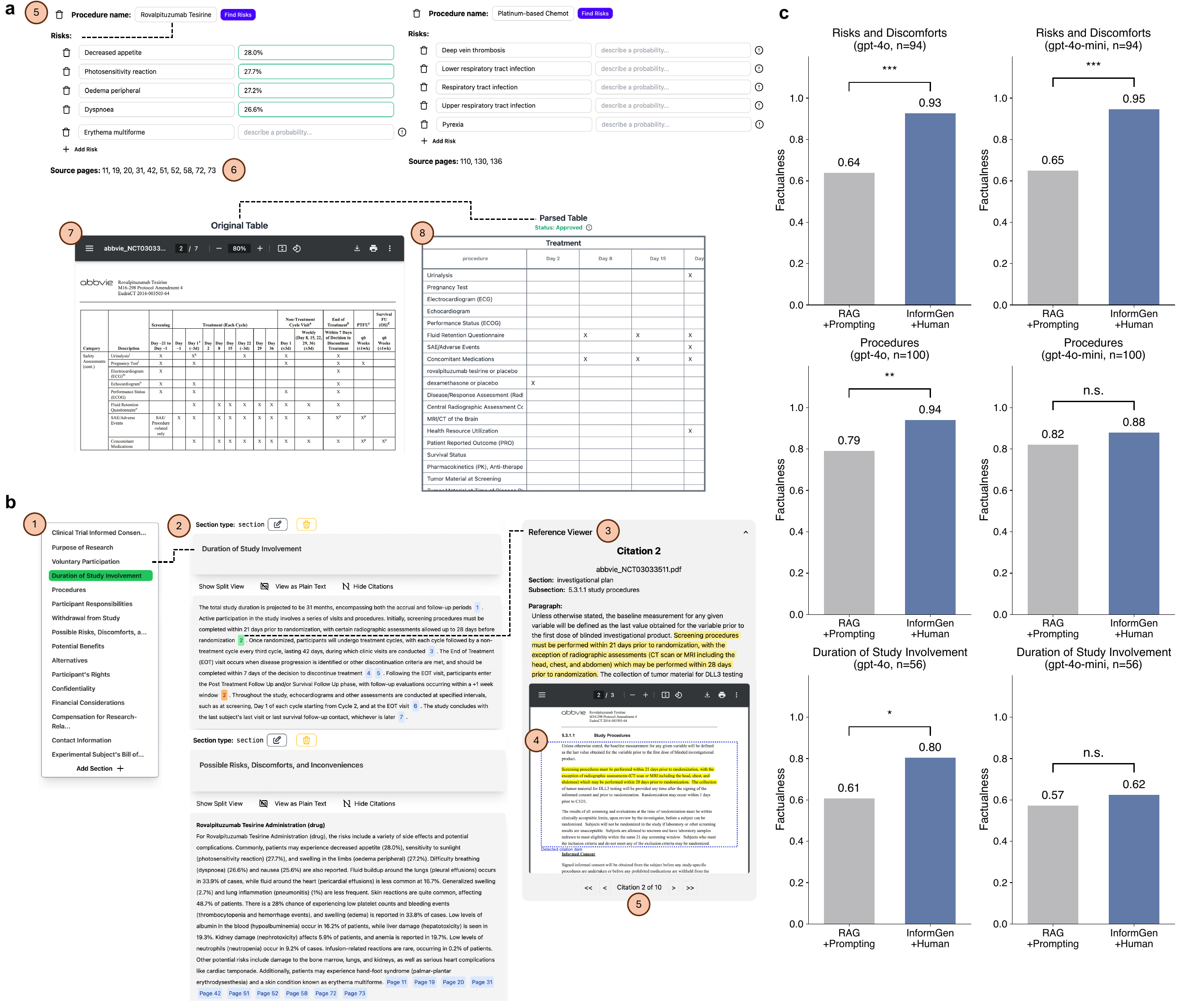}
    \caption{\textbf{Development of the \method platform for consent document generation with human-in-the-loop}. \textbf{a}, Illustration of the demo platform for AI-assisted consent document drafting. The interface allows users to review and refine the processed procedure-risk table and schedule of assessment (SOA) table extracted from the input protocol. Users can also trace back information to the original protocol for verification. \textbf{b}, Visualization of the generated ICF content, displayed section by section. The sidebar lists the generated sections, while the main panel presents the content with inline citations, which are linked to the source documents for user review. \textbf{c}, Manual evaluation of factual accuracy comparing RAG+Prompting with \method incorporating human oversight (\method+Human). The results show a significant accuracy improvement of 20\%-30\%, with \method+Human achieving over 90\% factual accuracy.
}
    \label{fig:manual_intervention}
\end{figure}

As illustrated in Figure~\hyperref[fig:overview]{1b}, \method utilizes the parsed consent template as input for site-specific policies and follows a structured pipeline for document generation: (1) generating a search query with metadata filters, (2) retrieving the most relevant chunks, and (3) drafting the section based on content guidance. In step (1), we leverage the table of contents from the target protocol, along with content guidance and relevant keywords, to prompt the LLM to synthesize a search query. This query includes metadata-based filtering to determine which sections of the protocol should be retrieved. This approach is particularly beneficial for lengthy protocols segmented into thousands of chunks, where a purely embedding-based ranking may overlook critical information. By incorporating metadata filtering, we significantly reduce the candidate set, improving retrieval precision and ensuring that essential content is effectively captured. In step (3), alongside generating content, we also instruct the model to produce inline citations linking to the retrieved input chunks. These citations enable users to trace the generated information back to the original protocol document for verification and review (Figure~\hyperref[fig:manual_intervention]{5b}).

\method incorporates specialized inputs for sections requiring precise details on schedules and risks associated with experimental procedures, specifically \textit{``Procedures"}, \textit{``Duration of Study Involvement"}, and \textit{``Risks and Discomforts"}. For the first two sections, the SOA table is used as an additional input. To facilitate processing, we segment the SOA table by time points, converting it into structured paragraphs that describe the activities occurring at each stage. These paragraphs are then provided to the model, which expands them into readable ICF content. Similarly, for the \textit{``Risks and Discomforts"} section, risks are categorized by procedure and transformed into paragraph form, allowing the LLM to generate well-structured and accurate content.



\bibliographystyle{naturemag}
\bibliography{main}

\clearpage

\clearpage

\appendix

\captionsetup[table]{name=Supplementary Table}
\setcounter{figure}{0}
\renewcommand*{\figurename}{Supplementary Figure}

\begin{figure}
    \centering
    \includegraphics[width=\linewidth]{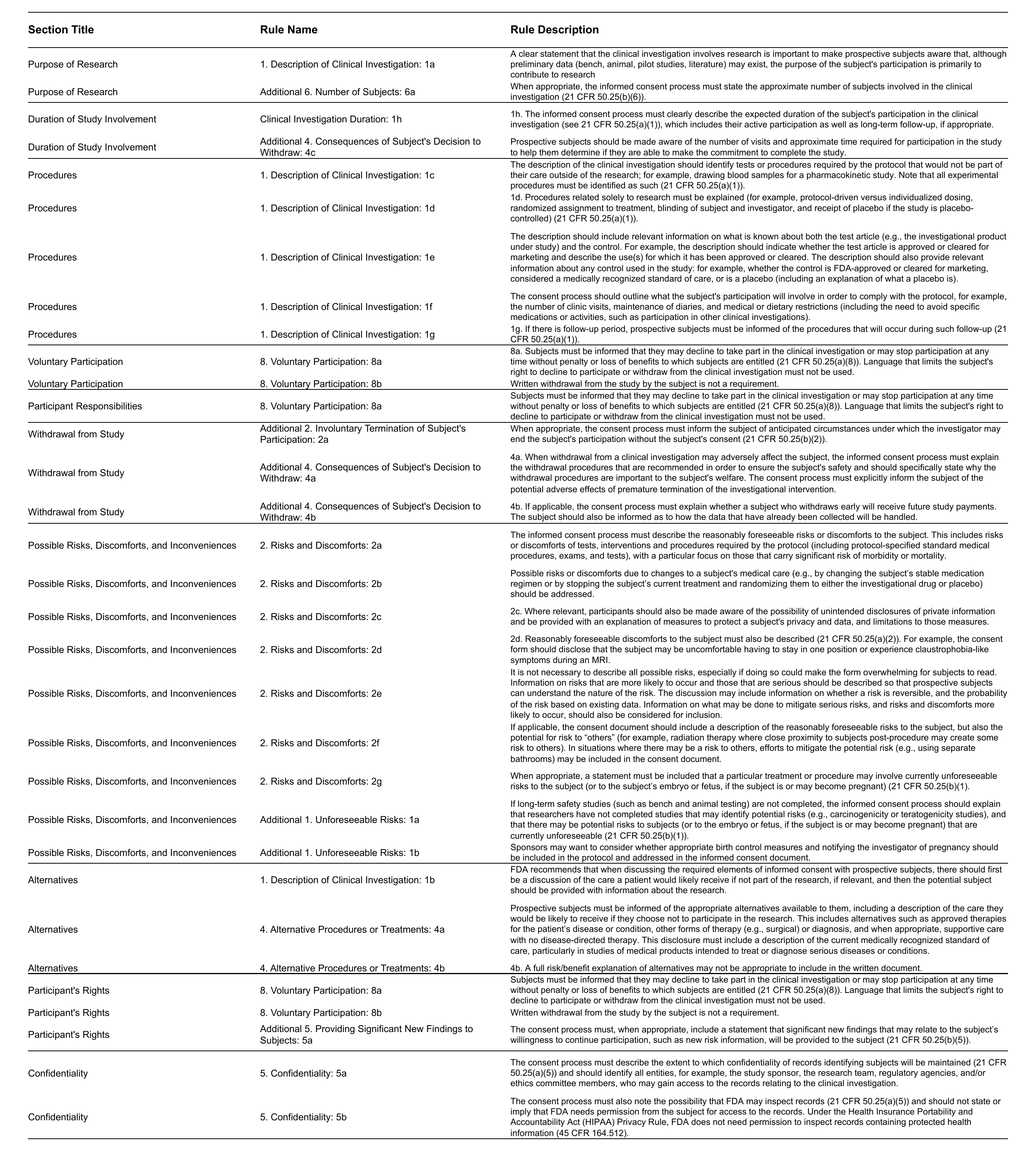}
    \caption{\textbf{Rule set for evaluating compliance of generated content.} \textit{``Section Title"}: Manually assigned section to which the rule applies and is used for assessment. \textit{``Rule Name"}: The designated name of the rule. \textit{``Rule Description"}: A detailed explanation of the rule, derived from the FDA guideline document.
}
    \label{fig:fda_policy}
\end{figure}

\end{document}